%% file: main.tex
\documentclass{article} 
\usepackage{iclr2026_conference, times}
\input{math_commands.tex}

\usepackage{hyperref}
\usepackage{url}
\usepackage{amsmath}
\usepackage{amssymb}
\usepackage{graphicx}
\usepackage{subcaption}
\usepackage{float}
\usepackage{geometry}
\usepackage[dvipsnames]{xcolor}
\usepackage{soul}
\sethlcolor{BurntOrange}
\usepackage[most]{tcolorbox}

\usepackage{amsthm}
\usepackage{algorithm}
\usepackage{algpseudocode}
\usepackage{bm}
\theoremstyle{remark}
\newtheorem{remark}{Remark}

\usepackage{parskip}
\usepackage{booktabs}
\usepackage{multirow}
\usepackage{marvosym}
\usepackage{setspace}

\title{Robust Mixture Models for Algorithmic Fairness Under Latent Heterogeneity}

\author{Siqi Li\thanks{Equal contribution.}
\hspace{1pt}\textsuperscript{\,$\dagger$} \\
Duke-NUS Medical School \\
\And
Molei Liu\footnotemark[1] \\
Peking University\\
\And
Yiwei Lyu \\
University of Michigan\\
\AND
Ziye Tian \\
Duke University\\
\And
Chuan Hong \\
Duke University\\
\And
Nan Liu \\
Duke-NUS Medical School \\
}

\iclrfinalcopy 

\begin{document}
\maketitle
\setstretch{1.25}
{\renewcommand{\thefootnote}{\ensuremath{\dagger}}\footnotetext[2]{siqili@u.duke.nus.edu}}

\begin{abstract}
Machine learning models optimized for average performance can perform poorly on vulnerable subpopulations. 
Existing approaches often rely on groups specified in advance, yet fairness-relevant subgroup structure may be latent, intersectional, and driven by complex interactions among continuous and discrete attributes.
We introduce \textbf{ROME} (\textbf{\underline{RO}}bust \textbf{\underline{M}}ixture \textbf{\underline{E}}nsemble), a framework that learns latent group structure while optimizing worst-group predictive performance. 
ROME connects latent-variable modeling with distributionally robust optimization (DRO) through two complementary approaches: an Expectation-Maximization formulation with robust aggregation for linear models and a neural Mixture-of-Experts formulation for nonlinear settings. 
Across simulations and three real-world regression datasets, ROME improves worst-group performance while maintaining competitive overall accuracy, including in comparisons with established group-aware and group-label-free robust learning methods. 
ROME provides a flexible approach to robust prediction when fairness-relevant attributes are available for subgroup discovery but their direct use in group-specific outcome models is restricted.
\end{abstract}

\vspace{-0.5em}

\section{Introduction}\label{sec.intro}
Machine learning models are increasingly used to support decisions in high-stakes domains such as healthcare~\citep{ning2024variable}, criminal justice~\citep{avila2020seductiveness}, and
finance~\citep{das2021fairness}.
Strong average predictive performance, however, does not ensure reliable performance for all subpopulations.
A model can achieve low overall error while performing substantially worse for smaller or systematically disadvantaged groups~\citep{barocas2016big}.
Identifying and reducing such disparities is therefore important for developing reliable and fair predictive systems.

A central difficulty is that vulnerable subpopulations may not be known in advance.
Even when sensitive attributes $\boldsymbol{S}$, such as sex, race, and socioeconomic characteristics, are available during model development, disparities may arise from unknown interactions among continuous and categorical attributes.
Manually defining such groups can require arbitrary discretization and produce many small intersectional groups~\citep{wu2025toward}.
At the same time, directly conditioning an outcome model $f(\cdot)$ on $\boldsymbol{S}$ may be undesirable in fairness-sensitive applications and may be restricted by institutional, policy, or legal requirements.
This creates an important tension because $\boldsymbol{S}$ can be valuable for identifying vulnerable subpopulations even when group-specific outcome models are restricted to non-sensitive predictive features $\boldsymbol{A}$.

\subsection{Related Work}
A large literature studies fairness and robustness when sensitive groups are observed and specified in advance. 
For regression, Fair Regression~\citep{pmlr-v97-agarwal19d} introduces bounded group loss constraints that control prediction loss within predefined groups, while GroupDRO~\citep{sagawa2020distributionallyrobustneuralnetworks} optimizes worst-group performance when group memberships are known.
These approaches provide strong protection for known groups but require the groups over which robustness is enforced to be specified in advance.

A complementary line of work seeks to protect vulnerable subpopulations without predefined demographic group labels.
ARL~\citep{ARL} adversarially reweights observations to emphasize poorly performing examples, while GEORGE~\citep{sohoni2020no} and JTT~\citep{liu2021just} identify hidden or error-prone subpopulations from learned representations or prediction errors.
The latter two were developed for classification, and extending their subgroup identification mechanisms to continuous outcomes requires additional choices that are not specified in the original methods.

Mixture models provide a classical framework for representing latent population heterogeneity~\citep{mclachlan2000finite,dempster1977maximum}, while mixture-of-experts models provide a flexible neural framework for learning heterogeneous predictive relationships
~\citep{jordan1994hierarchical,shazeer2017outrageously}.
Robust optimization provides complementary tools for protecting performance across heterogeneous populations
~\citep{sagawa2020distributionallyrobustneuralnetworks, wang2023distributionally}.

Taken together, existing work provides important tools when groups are predefined or demographic group information is unavailable.
We consider a complementary setting where $\boldsymbol{S}$ is available for learning subgroup structure, while group-specific outcome models depend only on $\boldsymbol{A}$.
We propose \textbf{ROME} (\textbf{\underline{RO}}bust \textbf{\underline{M}}ixture \textbf{\underline{E}}nsemble), which uses $\boldsymbol{S}$ to inform latent subgroup discovery and incorporates distributionally robust optimization (DRO) to improve performance for vulnerable groups, while restricting group-specific outcome models to non-sensitive features $\boldsymbol{A}$.

\subsection{Our Contributions}
First, we study group-robust prediction under latent population heterogeneity involving continuous and categorical fairness-relevant attributes.
Second, we develop two complementary formulations that connect latent-group modeling with robust optimization.
ROME-EM combines mixture regression with robust aggregation in linear settings, making explicit the connection between latent mixture components and robust source aggregation.
ROME-MoE extends this principle to nonlinear settings through a neural mixture-of-experts architecture that jointly learns soft subgroup structure and optimizes average and worst-group predictive performance.
Third, we evaluate ROME through simulation studies and three real-world regression datasets against five baselines spanning standard prediction, fair regression, robustness without predefined demographic group labels, and mixture-of-experts modeling.
We further examine performance across multiple external subgroup definitions and the trade-off between overall and worst-group predictive performance.

\section{Problem Setup}
We observe training data $\mathcal{D}=\{(\boldsymbol{A}_i,\boldsymbol{S}_i,Y_i)\}_{i=1}^n$, where $Y_i\in\mathbb{R}$ is a continuous outcome, $\boldsymbol{A}_i\in\mathbb{R}^{p_A}$ contains non-sensitive predictive features, and $\boldsymbol{S}_i\in\mathbb{R}^{p_S}$ contains fairness-relevant attributes.
We focus on settings in which $\boldsymbol{S}$ is available during model development to inform subgroup structure, while its direct use in group-specific outcome models may be restricted.
The population is assumed to contain $G$ latent subgroups whose memberships are not specified in advance.
Our goal is to learn this latent structure while improving predictive performance for poorly performing subpopulations without substantially sacrificing overall prediction accuracy.

\subsection{Group-Robust Fairness Objective}
We focus on disparities in predictive performance across subpopulations.
Let $\mathcal{G}$ denote a collection of subpopulations and let $Z \in \mathcal{G}$ denote group membership. 
For a predictor $f$, define the group-specific risk
\[
R_g(f)
=
\mathbb{E}\!\left[
\ell\{Y,f(\boldsymbol{X})\}
\mid Z=g
\right],
\qquad g\in\mathcal{G},
\]
where $\ell$ is the squared loss for the continuous outcomes considered in this work. 
Our primary group-robust criterion is the worst-group risk
\[
R_{\mathrm{worst}}(f)
=
\max_{g\in\mathcal{G}} R_g(f).
\]
Reducing $R_{\mathrm{worst}}(f)$ protects against disproportionately poor predictive performance in the most adversely affected subgroup, while overall risk measures population-level predictive accuracy. 
We therefore consider both worst-group and overall performance when evaluating the trade-off between subgroup robustness and predictive accuracy.

\begin{remark}
Our group-robust objective differs from classical parity-based fairness criteria. 
Demographic parity requires predictions to be independent of sensitive attributes, while equalized odds requires this independence conditional on the outcome~\citep{hardt2016equality}. 
Although extensions to continuous outcomes and sensitive attributes exist~\citep{pmlr-v97-agarwal19d,narasimhan2020pairwise}, they require formulations different from those commonly used in classification~\citep{bird2020fairlearn,hong2024trans,du2021fairness}. 
Our objective instead asks whether predictive error is disproportionately large for any subpopulation. 
Thus, low worst-group risk represents group-robust predictive performance and does not imply satisfaction of other fairness criteria.
\end{remark}

Importantly, the latent groups learned by ROME are distinct from the subgroups used for evaluation. 
ROME does not receive the evaluation partition as group labels during training. 
In the real-data experiments, externally defined intersectional groups are used only to audit subgroup performance and are not assumed to represent the latent groups learned by ROME.

\section{ROME-EM For Linear Models}\label{ROME-EM}
\subsection{Latent Mixture Model}\label{sec2.1}
ROME-EM models population heterogeneity through $G$ latent groups.
Let $z_{ij} \in \{0,1\}$ denote the latent group indicator where $z_{ij} = 1$ if observation $i$ belongs to group $j$. 
The membership probabilities follow a multinomial logistic model using selected sensitive attributes $\boldsymbol{S}_{i,\text{mem}} = \{S_{ik} : k \in \mathcal{I}_{\text{mem}}\}$
\begin{equation}\label{EM-assumption}
P(z_{ij} = 1 \mid \boldsymbol{S}_{i,\text{mem}}) = \frac{\exp(\gamma_j^\top \boldsymbol{S}_{i,\text{mem}})}{{\textstyle\sum_{k=1}^G} \exp(\gamma_k^\top \boldsymbol{S}_{i,\text{mem}})}
\end{equation}
where $\mathcal{I}_{\text{mem}} \subseteq \{1, \cdots, p_S\}$ denotes the indices of sensitive features used in the group membership model, and $\gamma_j \in \mathbb{R}^{|\mathcal{I}_{\text{mem}}|}$ are group-specific membership parameters.
Given group membership, the continuous outcome follows a group-specific linear model:
\begin{equation}
\label{formula:group-specific}
Y_i \mid z_{ij}=1 \sim \mathcal{N}(\boldsymbol{\omega}_j^\top \boldsymbol{X}_i, \sigma^2)
\end{equation}
where $\boldsymbol{X}_i = [1, \boldsymbol{A}_i, \boldsymbol{S}_{i,\text{out}}]$ is the design vector with $\boldsymbol{S}_{i,\text{out}} = \{S_{ik} : k \in \mathcal{I}_{\text{out}}\}$ denoting the sensitive features included in outcome prediction model and $\boldsymbol{\omega}_j \in \mathbb{R}^{1 + p_A + |\mathcal{I}_{\text{out}}|}$.
The feature selection indices $\mathcal{I}_{\text{mem}}$ and $\mathcal{I}_{\text{out}}$ enable flexible modeling based on domain constraints. For instance, we might use all demographic features to identify groups while excluding protected attributes from outcome prediction.
Throughout this work, we distinguish two roles of sensitive attributes:
$\boldsymbol{S}_{i,\text{mem}}$ contains variables allowed to inform the estimation of latent subgroup structure, while $\boldsymbol{S}_{i,\text{out}}$ contains variables permitted to enter the group-specific outcome model.\footnote{The determination of which attributes belong to $\boldsymbol{S}_{i,\text{mem}}$ and $\boldsymbol{S}_{i,\text{out}}$ is entirely guided by domain knowledge, institutional guidelines, and fairness/compliance requirements, not by the method itself.}

\subsection{EM Algorithm}\label{EM}
Since group memberships are latent, we use the expectation maximization (EM) algorithm to estimate parameters $\boldsymbol{\Theta} = \{\gamma_1, \ldots, \gamma_G, \boldsymbol{\omega}_1, \ldots, \boldsymbol{\omega}_G\}$.

\paragraph{Initialization:}
The initial hard assignments $\{z_{ij}^{(0)}\}_{i=1}^n$  provide a warm start for EM and are replaced by soft probabilities in subsequent iterations. They can either be 
(1) specified based on domain knowledge, for example by clustering sensitive attributes to obtain initial hard assignments, or
(2) randomly sampled from the discrete uniform distribution on $\{1, \ldots, G\}$.
For membership parameters $\gamma_j$, we fit a logistic regression of the initial group indicators on $\boldsymbol{S}_{\text{mem}}$. For outcome parameters $\boldsymbol{\omega}_j$, we fit group-specific linear regressions when $n_j$ is sufficiently large~\citep{ng2011algorithm,levine2001implementations}; otherwise, we use pooled regression estimates to ensure numerical stability.

\paragraph{E-Step:}
Given current parameters $\{\gamma_j^{(t)}, \boldsymbol{\omega}_j^{(t)}\}_{j=1}^G$, we compute the posterior probabilities:
\[
w_{ij}^{(t)} = \frac{p_{ij}^{(t)} \cdot \ell_{ij}^{(t)}}{\sum_{k=1}^G p_{ik}^{(t)} \cdot \ell_{ik}^{(t)}},
\]
where $p_{ij}^{(t)} = P(z_{ij} = 1 \mid \boldsymbol{S}_{i,\text{mem}}; \gamma_j^{(t)})$ is the membership probability of observation $i$ under group $j$ and
\[
\ell_{ij}^{(t)} = \exp\left(-\frac{1}{2}(Y_i - \boldsymbol{\omega}_j^{(t)\top}\boldsymbol{X}_i)^2\right)
\]
is the Gaussian likelihood, with fixed variance absorbed into the proportionality.

\paragraph{M-Step with Backtracking Line Search:}
To promote monotonic increase in the log-likelihood, we employ backtracking line search \citep{liang2021gradient} for parameter updates. For each parameter update from $\boldsymbol{\theta}^{(t)}$ to $\boldsymbol{\theta}^{(t+1)}$, we set:
$
\boldsymbol{\theta}^{(t+1)} = \boldsymbol{\theta}^{(t)} + \alpha(\boldsymbol{\theta}_{\text{new}} - \boldsymbol{\theta}^{(t)})
$ starting with $\alpha = 0.5$ and halving $\alpha$ until the likelihood improves or $\alpha < \tau_2$.
For each group, we update $\gamma_j$ via weighted logistic regression with weights $w_{ij}^{(t)}$, using a quasi-binomial family to handle potential overdispersion:
\[
\gamma_j^{\text{new}} = \arg\max_{\gamma} \sum_{i=1}^n w_{ij}^{(t)} \left[\gamma^\top \boldsymbol{S}_{i,\text{mem}} - \log\left(\sum_{k=1}^G \exp(\gamma_k^\top \boldsymbol{S}_{i,\text{mem}})\right)\right]
\]
We solve the weighted least squares problem:
$\boldsymbol{\omega}_j^{\text{new}} = (\boldsymbol{X}^\top \boldsymbol{W}_j^{(t)} \boldsymbol{X})^{-1} \boldsymbol{X}^\top \boldsymbol{W}_j^{(t)} \boldsymbol{Y}$,
where $\boldsymbol{W}_j^{(t)} = \text{diag}(w_{1j}^{(t)}, \ldots, w_{nj}^{(t)})$.

\paragraph{Convergence and Model Selection:}
The algorithm terminates when the total parameter change falls below tolerance $\tau_1$:
\[
\sum_{j=1}^G \left(||\gamma_j^{(t+1)} - \gamma_j^{(t)}||_1 + ||\boldsymbol{\omega}_j^{(t+1)} - \boldsymbol{\omega}_j^{(t)}||_1\right) < \tau_1
\]
or the maximum iterations are reached. 
As with finite mixture models generally, convergence need not correspond to a global optimum and may depend on initialization.
In practice, the number of groups $G$ can be selected using information criteria such as AIC or BIC computed from the maximized log-likelihood.

\subsection{DRO Aggregation}
\label{sec:dro-aggregation}
After obtaining group-specific models $\{\boldsymbol{\hat{\omega}}_1, \ldots, \boldsymbol{\hat{\omega}}_G\}$ from the EM, 
we treat the estimated latent components as source populations and apply the distributionally robust optimization (DRO) framework to construct a robust predictor that performs well across different mixture distributions of the latent groups. 

\paragraph{DRO Formulation.}
We consider mixture distributions of the $G$ group-specific conditionals:
\[
\mathbb{P}^q_{Y|X} = \sum_{j=1}^G q_j \,\mathbb{P}^{(j)}_{Y|X},
\qquad q \in \mathcal{H},
\]
where $\mathcal{H} = \left\{ \mathbf{v} \in \Delta^{G} : \|\mathbf{v} - \mathbf{v}_0\|_{2} \leq c \sqrt{G} \right\}\subseteq \Delta^G$ is an uncertainty set over mixture weights.
Here $\mathbf{v}_0$ represents a baseline mixture weight, taken by default as
the uniform vector $(1/G,\ldots,1/G)$, and $c\in[0,1]$ controls the size of the uncertainty region.
The distributionally robust predictor solves
\[
f^* = \arg\max_{f \in \mathcal{F}} \; \min_{q \in \mathcal{H}} R_q(f)
\]
where $R_q(f)$ denotes the explained variance under mixture weights $q$.

\paragraph{Closed-Form Solution:} 
Following key results from~\citet{wang2023distributionally}, when $\mathcal{F}$ consists of convex combinations of the group-specific estimated predictors $\hat{f}_j(\boldsymbol{X}) = \hat{\boldsymbol{\omega}}_j^{\top}\boldsymbol{X}$, the robust predictor $f_{\boldsymbol{v}}$ has the closed form: 
\begin{equation}
    f_{\boldsymbol{v}}(\boldsymbol{X}_i) = \sum_{j=1}^G v_j \cdot \hat{\boldsymbol{\omega}}_j^{\top}\boldsymbol{X}_i
\end{equation}
where the optimal weights solve:
\begin{equation}\label{v}
    \boldsymbol{v}^* = \arg\min_{v \in \mathcal{H}} \boldsymbol{v}^{\top} \hat{\gamma}\boldsymbol{v}
\end{equation}
Here $\gamma$ is a $G \times G$ matrix, representing the empirical covariance of predictions across groups, with entries $\gamma_{k,j} = \mathbb{E}_{Q_X}[f^{(k)}(X)f^{(j)}(X)]$ and we compute $\hat{\gamma}$ by $\hat{\gamma}_{jk} = \frac{1}{n} \sum_{i=1}^n \hat{f}_j(\boldsymbol{X}_i) \cdot \hat{f}_k(\boldsymbol{X}_i) $. 

\paragraph{Final Robust Predictor:} Finally, the robust predictor of ROME-EM for $\boldsymbol{X}_{\text{new}}$ is: 
\begin{equation}
    \hat{Y}_{\text{ROME-EM}} = \sum_{j=1}^G v_j^* \cdot \hat{\boldsymbol{\omega}}_j^{\top}\boldsymbol{X}_{\text{new}}
\end{equation}

ROME-EM makes explicit the connection between latent mixture estimation and robust aggregation. 
This linear construction provides the methodological motivation for extending latent-group robustness to the nonlinear mixture-of-experts setting developed next.

\section{ROME-MoE for Nonlinear Models}\label{MoE}
\begin{algorithm}[t]
\caption{ROME-MoE}
\label{alg:rome-moe}
\begin{algorithmic}[1]
\Require Training data $\mathcal{D} = \{(\boldsymbol{A}_i, \boldsymbol{S}_i, Y_i)\}_{i=1}^n$, number of experts $G$, DRO parameter $\alpha \in [0,1]$, learning rate $\eta$, batch size $B$, epochs $E$
\Ensure Trained ROME-MoE model with parameters $\Theta = \{\theta_{\text{gate}}, \theta_1, \ldots, \theta_G\}$

\State \textbf{Initialize:} Gating network $g_{\theta_{\text{gate}}}$, Expert networks $\{f_{\theta_1}, \ldots, f_{\theta_G}\}$

\For{epoch $= 1$ to $E$}
    \For{each batch $\mathcal{B} \subset \mathcal{D}$ with $|\mathcal{B}| = B$}
        \For{each $(\boldsymbol{A}_i, \boldsymbol{S}_i, Y_i) \in \mathcal{B}$}
            \If{\texttt{ROME-MoE-S}}
                \State $\boldsymbol{w}_i = \text{Softmax}(g_{\theta_{\text{gate}}}(\boldsymbol{S}_i))$ \Comment{\texttt{ROME-MoE-S}}
            \Else \Comment{\texttt{ROME-MoE-AS}}
                \State $\boldsymbol{w}_i = \text{Softmax}(g_{\theta_{\text{gate}}}([\boldsymbol{A}_i; \boldsymbol{S}_i]))$
            \EndIf
            \State $\hat{Y}_i = \sum_{j=1}^G w_{ij} \cdot f_{\theta_j}(\boldsymbol{A}_i)$ \Comment{Weighted expert predictions}
        \EndFor
        
        \For{$j = 1$ to $G$}
            \State $\mathcal{I}_j = \{i \in \mathcal{B} : w_{ij} > 0.1\}$ \Comment{Samples with significant membership}
            \If{$|\mathcal{I}_j| > 0$}
                \State $\mathcal{L}_j = \frac{1}{|\mathcal{I}_j|} \sum_{i \in \mathcal{I}_j} w_{ij} \cdot (Y_i - \hat{Y}_i)^2$
            \Else
                \State $\mathcal{L}_j = 0$
            \EndIf
        \EndFor
        
        \State \textbf{// DRO Objective}
        \State $\mathcal{L}_{\text{avg}} = \frac{1}{B} \sum_{i \in \mathcal{B}} (Y_i - \hat{Y}_i)^2$ \Comment{Average loss}
        \State $\mathcal{L}_{\text{worst}} = \max_{j \in \{1,\ldots,G\}} \mathcal{L}_j$ \Comment{Worst group loss}
        \State $\mathcal{L}_{\text{total}} = (1-\alpha) \cdot \mathcal{L}_{\text{avg}} + \alpha \cdot \mathcal{L}_{\text{worst}}$
        
        \State \textbf{// Backward Pass}
        \State Compute gradients: $\nabla_{\Theta} \mathcal{L}_{\text{total}}$
        \State Update parameters: $\Theta \leftarrow \Theta - \eta \cdot \nabla_{\Theta} \mathcal{L}_{\text{total}}$
    \EndFor
\EndFor

\State \Return Trained model with parameters $\Theta$
\end{algorithmic}
\end{algorithm}

While ROME-EM follows a classical EM-style update structure and includes a closed-form robust aggregation step under mixture model assumptions, the linear multinomial assumption for group membership (Equation~\ref{EM-assumption}) makes ROME-EM vulnerable to misspecification when true group structures involve non-linear or complex feature interactions.

Although neural network variants of mixture models have been developed~\citep{nagpal2022deepcoxmixturessurvival}, they do not retain the simple
closed-form robust aggregation structure used by ROME-EM.
We therefore pursue a different approach: extending ROME to neural networks using Mixture of Experts (MoE)~\citep{shazeer2017outrageously}, which naturally handles non-linear relationships while incorporating DRO principles directly into the training objective.
Specifically, ROME-MoE treats experts as group-specific predictors and uses a loss function that explicitly balances average and worst-group performance, operationalizing the maximin principle from Section~\ref{sec:dro-aggregation} in an end-to-end differentiable framework.

Our ROME-MoE architecture consists of two key components: 
(1) \textbf{Gating network:} A neural network $g: \mathbb{R}^{p} \rightarrow \mathbb{R}^{G}$ maps input features to $G$ logits, which are transformed by a softmax to obtain a probability distribution over the experts and determine soft group assignments.
(2) \textbf{Expert network:} A collection of $G$ neural networks $\{f_1, \ldots, f_G\}$ where each $f_j: \mathbb{R}^{p_A} \rightarrow \mathbb{R}$ specializes in predictions for its corresponding latent group using only non-sensitive features.
We present two variants: ROME-MoE-S uses only sensitive features $\boldsymbol{S}$ for the gating network (assuming group membership is primarily determined by $\boldsymbol{S}$), while ROME-MoE-AS uses all features. 
The key architectural restriction in ROME-MoE is that $\boldsymbol{S}$ can influence group assignment through the gating network but is not provided directly to the expert prediction functions, which receive only $\boldsymbol{A}$.

\paragraph{Use of sensitive attributes.}
Sensitive features $\boldsymbol{S}$ are used by the gating network to inform latent subgroup structure, while the expert outcome models receive only $\boldsymbol{A}$. 
Thus, $\boldsymbol{S}$ can influence the final prediction through the gating weights but does not enter the expert prediction functions directly. 
Users determine which attributes are treated as sensitive according to the domain and relevant institutional, policy, or legal requirements.

The detailed implementation is available in Algorithm \ref{alg:rome-moe}. Unlike Vanilla MoE training that minimizes average loss, ROME-MoE incorporates distributionally robust optimization directly into the training objective. Given a batch of data, we compute group assignments through the gating network and optimize:
\begin{equation}
\mathcal{L}_{\text{ROME-MoE}} = (1-\alpha) \cdot \mathcal{L}_{\text{avg}} + \alpha \cdot \mathcal{L}_{\text{worst}}
\end{equation}
where $\mathcal{L}_{\text{avg}}$ is the average loss across all samples, $\mathcal{L}_{\text{worst}} = \max_{j \in \{1, \ldots, G\}}\mathcal{L}_j$ is the worst group loss, and $\alpha \in [0,1]$ controls the trade-off between average and worst-case performance.  
For each latent group, $\mathcal{L}_j$ is computed using observations with membership weight $w_{ij}>0.1$, with their squared errors weighted by $w_{ij}$, as detailed in Algorithm~\ref{alg:rome-moe}.
In practice, we found $\alpha \in [0.05, 0.1]$ provides a good balance between average and worst-group performance. 

\section{Experiments}
\subsection{Simulation Study}\label{simulations}
\paragraph{Data Generation and Setup}
We conducted a comprehensive simulation study to evaluate ROME-EM's ability to recover group-specific parameters and improve worst-group performance. 
We generated data with $n=8,000$ observations from $G=4$ latent groups, each with distinct parameter vectors and mixing proportions. 
Group membership was determined by a multinomial logistic model based on sensitive attributes $\boldsymbol{S} \in \mathbb{R}^5$, with each group having distinct membership parameters. For the outcome model, we generated $p=20$ total features comprising $p_A=15$ non-sensitive features and $p_S=5$ sensitive attributes. 

The true regression coefficients $\beta_j$ for each group were constructed using a scaled perturbation approach to introduce meaningful heterogeneity across groups while maintaining realistic parameter magnitudes. 
Complete simulation parameters are provided in Appendix~\ref{app:sim_para}, with additional details in Appendix~\ref{syn_details}.
To assess robustness to initialization, we intentionally misspecified the starting conditions by randomly reassigning $50\%$ of observations to incorrect groups before running ROME-EM. 

For the DRO optimization stage, we evaluated multiple constraint values to examine the trade-off between overall and worst-group predictive performance (detailed in Appendix~\ref{app:sim_para}).
Each of the 100 simulation replications was evaluated on an independent test set of $n=8,000$ observations generated from the same underlying distributions.

\begin{figure}[!htbp]
    \centering
    \includegraphics[width=1\linewidth]{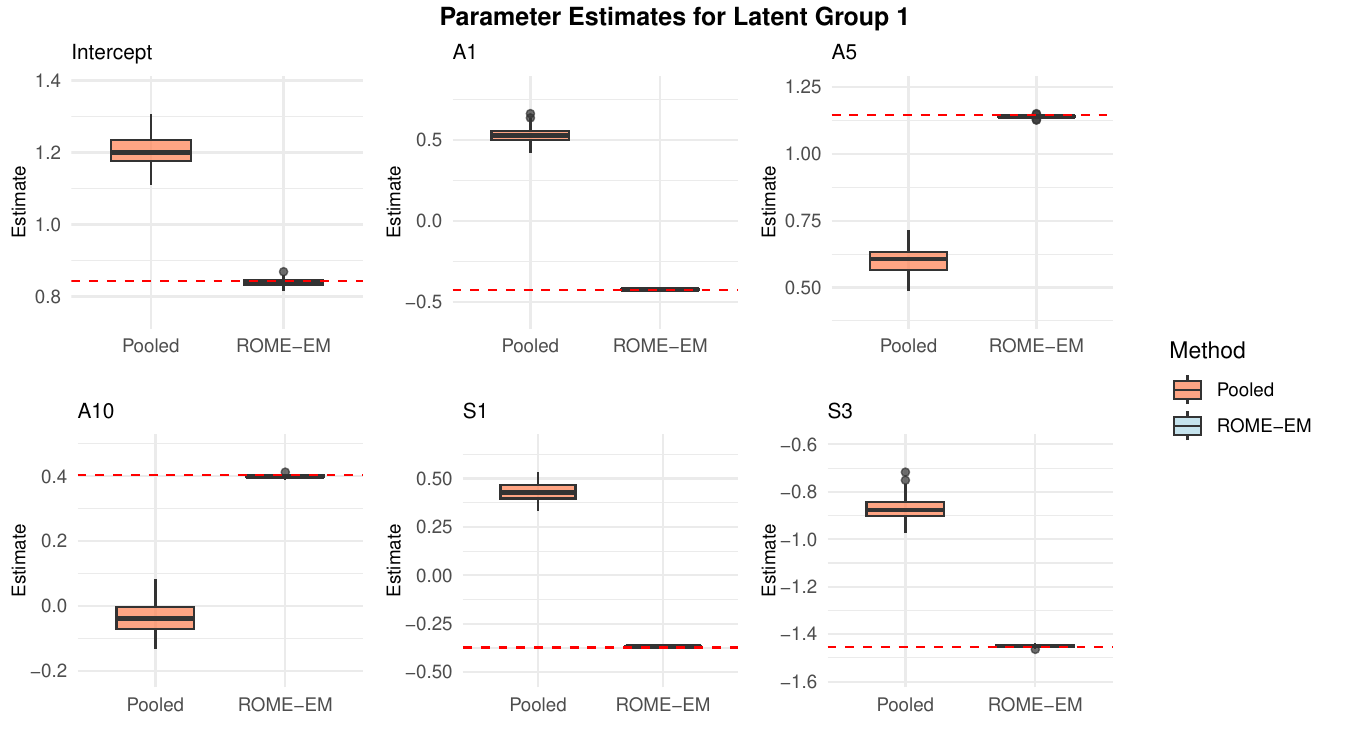}
    \caption{Box plots comparing parameter recovery by ROME-EM and pooled regression for latent group 1 (over 100 simulations). Red dashed lines denote ground truth values for each parameter.}
    \label{fig:bias_group1}
\end{figure}

\begin{figure}[!htbp]
    \centering
    \includegraphics[width=0.8\linewidth]{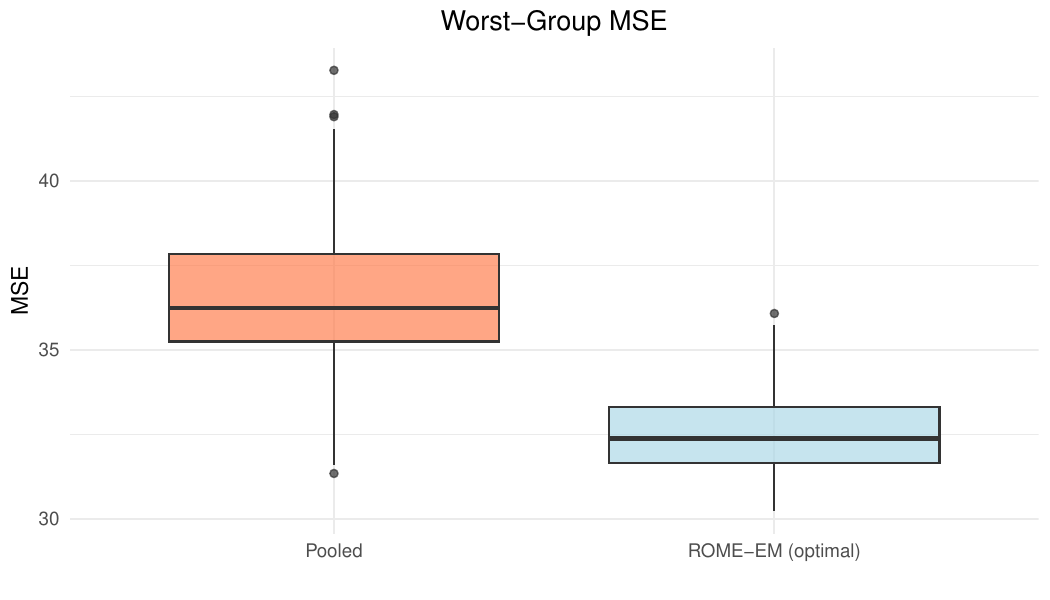}
    \caption{Box plots comparing worst-group MSE over 100 simulations.}
    \label{fig:mse_sim}
\end{figure}

\paragraph{Results}
Across 100 simulation replications, ROME-EM demonstrated superior parameter recovery and worst-group performance compared to pooled regression.
Figure \ref{fig:bias_group1} illustrates the parameter estimation accuracy for a representative latent group, showing that ROME-EM estimates (blue) consistently cluster near the true values (red dashed lines) while pooled estimates (coral) exhibit substantial bias. 
The remaining results for the other three latent groups are available in Appendix \ref{app:additional_groups} and show similar patterns.
The results also show a statistically significant improvement in worst-group performance with ROME-EM.
As shown in Figure \ref{fig:mse_sim}, the worst-group MSE decreased from a mean of $36.56$ under pooled regression to a mean $32.59$ with ROME-EM (optimal constraint), representing an approximately $11\%$ reduction (paired t-test: $p < 0.001$). 

\subsection{Real-Data Analysis}\label{realdata}

\begin{table}[t]
\centering
\caption{Overall and worst-group Mean Squared Error (MSE) on the test sets, averaged over 10 random seeds (mean $\pm$ standard error).
Best-performing fair models are shown in \textbf{bold}.
Statistical significance is assessed using paired two-sided t-tests across the 10 matched random seeds.
For worst-group MSE, each ROME variant is compared with the best-performing fair baseline.
For overall MSE, each ROME variant is compared with the best-performing baseline among all models.
Statistical significance is denoted by $^*p<0.05$, $^{**}p<0.01$, and $^{***}p<0.001$; $^{\mathrm{ns}}$ indicates no significant difference ($p>0.05$).}
\label{tab:main_results_mse}
\begin{tabular}{@{}llcll@{}}
\toprule
\textbf{Dataset} & \textbf{Model} & \textbf{Fair} & \textbf{Overall MSE}  & \textbf{Worst-Group MSE} \\ \midrule
\multirow{3}{*}{\begin{tabular}[c]{@{}l@{}} Law School Admissions \\ Council \end{tabular}} 
 & Baseline MLP &  & 0.7196 $\pm$ 0.0006 & 0.8184 $\pm$ 0.0027\\
 & Baseline MLP - Fair & \checkmark & 0.7368 $\pm$ 0.0005 & 0.8300 $\pm$ 0.0012 \\

& ARL & \checkmark & 0.7355 $\pm$ 0.0004 & 0.8287 $\pm$ 0.0021 \\

& FairReg-BGL & \checkmark & 0.7352 $\pm$ 0.0005 & 0.8269 $\pm$ 0.0017 \\
 
 & Vanilla MoE &  & 0.7209 $\pm$ 0.0007 & 
 0.8233 $\pm$ 0.0018 \\
 & ROME-MoE-S & \checkmark & 0.7217 $\pm$ 0.0014$^{\mathrm{ns}}$ & 
0.8144 $\pm$ 0.0011$^{***}$ \\
 & ROME-MoE-AS & \checkmark & \textbf{0.7200 $\pm$ 0.0015}$^{\mathrm{ns}}$ & 
\textbf{0.8127 $\pm$ 0.0023}$^{***}$ \\ 
 \midrule

\multirow{3}{*}{Communities and Crime} 
 & Baseline MLP &  & 0.0205 $\pm$ 0.0002 & 0.0285 $\pm$ 0.0005 \\ 
 & Baseline MLP - Fair & \checkmark & 0.0235 $\pm$ 0.0002 & 0.0326 $\pm$ 0.0004 \\
  
 & ARL & \checkmark & 0.0225 $\pm$ 0.0001 & 0.0300 $\pm$ 0.0003 \\

 & FairReg-BGL & \checkmark & 0.0223 $\pm$ 0.0001 & 0.0298 $\pm$ 0.0002 \\
 
 & Vanilla MoE &  & 0.0207 $\pm$ 0.0002 & 
 0.0286 $\pm$ 0.0005 \\
 & ROME-MoE-S & \checkmark & 0.0207 $\pm$ 0.0002$^{\mathrm{ns}}$  & 
 0.0287 $\pm$ 0.0004$^{**}$ \\
 & ROME-MoE-AS & \checkmark & \textbf{0.0204 $\pm$ 0.0001}$^{\mathrm{ns}}$ & 
\textbf{0.0271 $\pm$ 0.0002}$^{***}$ \\
 \midrule
 
\multirow{3}{*}{\begin{tabular}[c]{@{}l@{}} American Community Survey \\ Public Use Microdata Sample \end{tabular}} 
 & Baseline MLP &  & 0.0044 $\pm$ 0.0000 & 0.0046 $\pm$ 0.0000 \\
 & Baseline MLP - Fair & \checkmark & 0.0053 $\pm$ 0.0000 & 0.0053 $\pm$ 0.0000 \\

 & ARL & \checkmark & 0.0052 $\pm$ 0.0000 & 0.0053 $\pm$ 0.0000 \\

 & FairReg-BGL & \checkmark & 0.0053 $\pm$ 0.0000 & 0.0053 $\pm$ 0.0000 \\
 
 & Vanilla MoE &  & 0.0047 $\pm$ 0.0000 & 
0.0048 $\pm$ 0.0000 \\

 & ROME-MoE-S & \checkmark & \textbf{0.0045 $\pm$ 0.0001}$^{\mathrm{ns}}$ & 
 \textbf{0.0047 $\pm$ 0.0001}$^{***}$ \\
 & ROME-MoE-AS & \checkmark & 0.0047 $\pm$ 0.0001 & 
 0.0049 $\pm$ 0.0000$^{***}$ \\
 \bottomrule
\end{tabular}
\end{table}

\begin{table}[t]
\centering
\caption{Overall and worst-group $R^2$ on the test sets, averaged over 10 random seeds (mean $\pm$ standard error). 
Best-performing fair models are
shown in \textbf{bold}. 
Statistical significance is assessed using paired two-sided t-tests across the 10 matched random seeds.
For worst-group $R^2$, each ROME variant is compared with the best-performing fair baseline. 
For overall $R^2$, each ROME variant is compared with the best-performing baseline among all models. 
Statistical significance is denoted by $^*p<0.05$, $^{**}p<0.01$, and $^{***}p<0.001$; $^{\mathrm{ns}}$ indicates no significant difference ($p>0.05$).}
\label{tab:main_results_r2}
\begin{tabular}{@{}llcll@{}}
\toprule
\textbf{Dataset} & \textbf{Model} & \textbf{Fair} & \textbf{Overall \boldsymbol{$R^2$}}  & \textbf{Worst-Group \boldsymbol{$R^2$}} \\ \midrule
\multirow{3}{*}{\begin{tabular}[c]{@{}l@{}} Law School Admissions \\ Council \end{tabular}} 
 & Baseline MLP &  & 0.1776 $\pm$ 0.0007 & 0.1225 $\pm$ 0.0006 \\
 & Baseline MLP - Fair & \checkmark & 0.1579 $\pm$ 0.0005 & 0.0933 $\pm$ 0.0015 \\

& ARL & \checkmark & 0.1594 $\pm$ 0.0004 & 0.0929 $\pm$ 0.0017 \\

& FairReg-BGL & \checkmark & 0.1597 $\pm$ 0.0005 & 0.0928 $\pm$ 0.0008 \\
 
 & Vanilla MoE &  & 0.1761 $\pm$ 0.0008 & 
0.1210 $\pm$ 0.0012 \\
 & ROME-MoE-S & \checkmark & 0.1752 $\pm$ 0.0016$^{\mathrm{ns}}$ & 
0.1148 $\pm$ 0.0024$^{***}$ \\
 & ROME-MoE-AS & \checkmark & \textbf{0.1772 $\pm$ 0.0017}$^{\mathrm{ns}}$ & 
\textbf{0.1181 $\pm$ 0.0021}$^{***}$ \\ \midrule

\multirow{3}{*}{Communities and Crime} 
 & Baseline MLP &  & 0.6149 $\pm$ 0.0029 & 0.5274 $\pm$ 0.0025 \\
 & Baseline MLP - Fair & \checkmark & 0.5570 $\pm$ 0.0032 & 0.4882 $\pm$ 0.0062 \\

& ARL & \checkmark & 0.5751 $\pm$ 0.0019 & 0.5276 $\pm$ 0.0045 \\

& FairReg-BGL & \checkmark & 0.5793 $\pm$ 0.0016 & \textbf{0.5361 $\pm$ 0.0018} \\
  
 & Vanilla MoE &  & 0.6105 $\pm$ 0.0033 & 
 0.5281 $\pm$ 0.0030 \\
 & ROME-MoE-S & \checkmark & 0.6095 $\pm$ 0.0032$^{\mathrm{ns}}$  & 
0.5270 $\pm$ 0.0031 \\
 & ROME-MoE-AS & \checkmark & \textbf{0.6166 $\pm$ 0.0024}$^{\mathrm{ns}}$ & 
 0.5356 $\pm$ 0.0036$^{\mathrm{ns}}$ \\ \midrule
 
\multirow{3}{*}{\begin{tabular}[c]{@{}l@{}} American Community Survey \\ Public Use Microdata Sample \end{tabular}} 
 & Baseline MLP &  & 0.5463 $\pm$ 0.0035 & 0.4907 $\pm$ 0.0047 \\
 & Baseline MLP - Fair & \checkmark & 0.4554 $\pm$ 0.0024 & 0.4237 $\pm$ 0.0031 \\
 
 & ARL & \checkmark & 0.4571 $\pm$ 0.0020 & 0.4245 $\pm$ 0.0036 \\

& FairReg-BGL & \checkmark & 0.4543 $\pm$ 0.0032 & 0.4272 $\pm$ 0.0024 \\
 
 & Vanilla MoE &  & 0.5104 $\pm$ 0.0034 & 
 0.4663 $\pm$ 0.0036 \\
 & ROME-MoE-S & \checkmark & \textbf{0.5299 $\pm$ 0.0078}$^{\mathrm{ns}}$ & 
 \textbf{0.4779 $\pm$ 0.0093}$^{**}$ \\
 & ROME-MoE-AS & \checkmark & 0.5170 $\pm$ 0.0054 & 
 0.4554 $\pm$ 0.0053$^{**}$ \\
 \bottomrule
\end{tabular}
\end{table}

\paragraph{Results}
We evaluated our approach on three real-world datasets from diverse domains: the Law School Admissions dataset~\citep{wightman1999lsac}, the Communities \& Crime dataset~\citep{communities_and_crime_183}, and the American Community Survey Public Use Microdata Sample (ACS PUMS)~\citep{us2003american}, accessed via the Folktables library~\citep{ding2021retiring}.

For each dataset, we compared seven models: 
(1) Baseline MLP, a standard MLP using both $\boldsymbol{A}$ and $\boldsymbol{S}$ as predictive features;
(2) Baseline MLP-Fair, an MLP using only non-sensitive features $\boldsymbol{A}$;
(3) Adversarially Reweighted Learning (ARL)~\citep{ARL}, which adversarially reweights poorly performing observations without requiring predefined demographic groups and is adapted here to continuous outcomes using squared error;
(4) Fair Regression-BGL~\citep{pmlr-v97-agarwal19d}, which minimizes overall prediction loss subject to bounded loss within predefined sensitive groups;
(5) Vanilla MoE, which uses both $\boldsymbol{A}$ and $\boldsymbol{S}$ for gating and expert prediction;
(6) ROME-MoE-S, which uses $\boldsymbol{S}$ for latent-group gating and $\boldsymbol{A}$ for expert prediction; and
(7) ROME-MoE-AS, which uses both $\boldsymbol{A}$ and $\boldsymbol{S}$ for latent-group gating while restricting expert prediction to $\boldsymbol{A}$.
Under the architectural criterion used in our experiments, models (2), (3), (4), (6), and (7) are considered as fair models because sensitive attributes are not provided as direct inputs to their outcome prediction models.

We set the DRO parameter $\alpha=0.05$ for all main ROME-MoE experiments and examine its effect in the ablation study in Section~\ref{sec:ablation}.
We report both MSE and $R^2$ for overall test-set performance and for the worst-performing evaluation subgroup.
Because ground-truth latent groups are unavailable in the real datasets, worst-group performance is audited using externally defined subgroups based on sensitive attributes. 
The main evaluation uses $(S_2,S_3)$ with quartile discretization for Law School, $(S_2,S_3)$ with median discretization for Communities \& Crime, and $S_2$ for ACS PUMS. 
Additional subgroup definitions are examined in the ablation study.

Tables~\ref{tab:main_results_mse} and~\ref{tab:main_results_r2} summarize the main results. ROME-MoE achieves strong worst-group performance while retaining competitive overall predictive accuracy. 
For worst-group MSE, both ROME variants significantly outperform the best-performing fair baseline on all three datasets. 
ROME-MoE-AS attains the lowest worst-group MSE among fair models on Law School and Communities \& Crime, while ROME-MoE-S attains the lowest worst-group MSE on ACS PUMS. 
At the same time, the overall MSE of the best-performing ROME variant on each dataset is not significantly different from that of the best-performing baseline among all models.

The $R^2$ results provide a complementary view. On Law School and ACS PUMS, ROME achieves the highest worst-group $R^2$ among fair models, with significant improvements over the best-performing fair baselines. 
On Communities \& Crime, Fair Regression-BGL attains the highest mean worst-group $R^2$ at $0.5361$, while ROME-MoE-AS achieves a nearly identical value of $0.5356$ with no significant difference between the two methods. 
ROME-MoE-AS nevertheless achieves the highest overall $R^2$ among the fair models on this dataset.
Together with the MSE results, these findings indicate that ROME improves worst-group prediction error while generally preserving strong overall predictive performance.

ROME-EM performed well in the simulation study but encountered difficulties on these real datasets. 
Without informative initialization or sufficiently strong mixture structure, the algorithm sometimes converged to solutions with similar parameters across groups, effectively approaching pooled regression, or failed to converge. 
These results suggest that the parametric mixture assumptions of ROME-EM can be restrictive for complex real-world heterogeneity. 
In contrast, ROME-MoE learns the latent structure and predictive models jointly through the gating and expert networks and was more effective in these real-data settings.

The main tables (\ref{tab:main_results_mse} and \ref{tab:main_results_r2})  report one representative evaluation subgroup scheme for each dataset. 
We additionally evaluate two alternative subgroup definitions per dataset in Section~\ref{sec:ablation}, with full results provided in Appendix~\ref{append.ablation.subgroups}.
Experimental details, including data preprocessing and hyperparameter selection, are provided in Appendices~\ref{realdatadetails} and~\ref{sec:appendix_hyperparams}.

\paragraph{Ablation Studies}\label{sec:ablation}
We conducted two ablation studies to examine key design choices in ROME.

\begin{figure}[!htbp]
    \centering
    \includegraphics[width=0.9\linewidth]{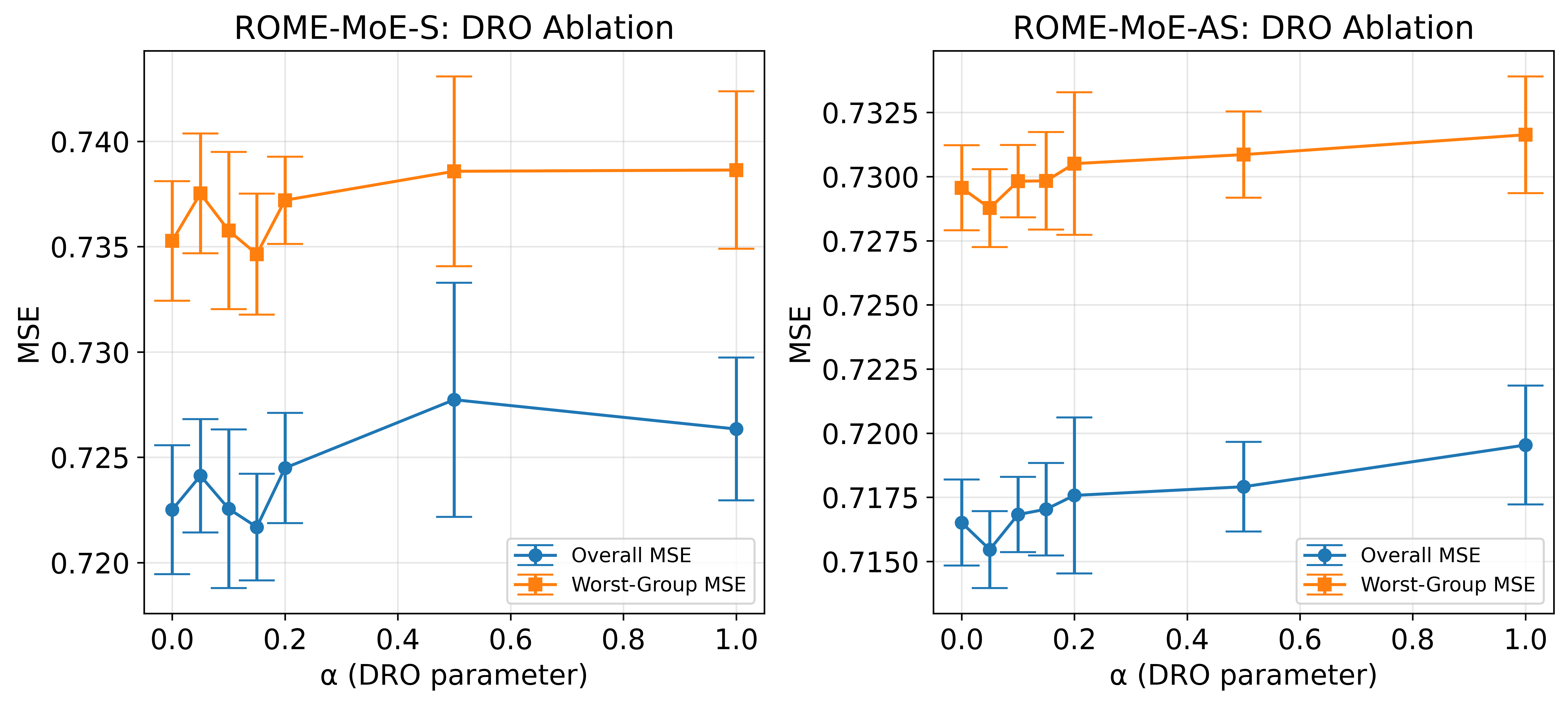}
    \caption{Effect of the DRO parameter $\alpha$ on ROME-MoE performance for the Law School Admissions Council dataset over 10 random seeds. 
    Both variants show improvements in worst-group performance while overall performance remains relatively stable across a broad range of $\alpha$, illustrating the trade-off controlled by the DRO objective.}
    \label{law}
\end{figure}

\textbf{1. Impact of DRO Parameter $\alpha$.} 
We varied the DRO parameter $\alpha$ to examine the contribution of the robust training objective beyond the MoE architecture alone.
As shown in Figures~\ref{crime}, \ref{law}, and~\ref{acs}, increasing $\alpha$ generally improves worst-group performance, with the gains tending to saturate at modest values of $\alpha$, while overall MSE remains relatively stable across a broad range of values.
These results support our choice of $\alpha=0.05$ for the main experiments, which places substantially greater weight on overall predictive accuracy while still incorporating worst-group robustness.
The broadly similar patterns across datasets and both ROME-MoE variants support the contribution of the DRO objective beyond the MoE architecture alone.

\textbf{2. Impact of Evaluation Subgroups:} 
To assess whether the observed improvements depend on a particular evaluation partition, we evaluated the models under multiple subgroup definitions based on different combinations of sensitive attributes.
Categorical attributes retain their natural categories, while continuous attributes are partitioned using median or quartile cut points.
Results for the alternative evaluation schemes are reported in Appendix~\ref{append.ablation.subgroups}.
Across these subgroup definitions, ROME generally maintains strong worst-group performance, indicating that the main findings are not driven by a single choice of evaluation partition.

\section{Discussion}\label{discussion}
ROME addresses group-robust prediction in settings where fairness-relevant attributes are observed but the subgroup structure associated with predictive disparities is not known in advance. 
Our results highlight complementary roles for the two formulations. 
ROME-EM provides an explicit connection between latent mixture modeling and robust aggregation and performs well when the mixture structure is correctly specified, as demonstrated in our simulations.
On the real datasets, however, it was sensitive to initialization and parametric model assumptions, sometimes producing similar group-specific solutions or failing to converge. 
ROME-MoE provides a more flexible nonlinear alternative and achieved strong worst-group performance across the three real-world datasets while generally maintaining competitive overall predictive performance.

An important feature of ROME is the separation between subgroup modeling and group-specific outcome prediction. 
Rather than requiring the relevant intersectional groups to be specified in advance, ROME uses fairness-relevant attributes to learn a compact soft representation of population heterogeneity.
In ROME-MoE, sensitive attributes may inform the gating network, while the expert outcome models receive only non-sensitive features. 
The learned latent groups are not themselves assumed to correspond to demographic categories.
Instead, performance can subsequently be audited using externally defined subgroups, as in our real-data experiments. 
This distinction allows latent structure to guide robust model fitting without treating any single predefined partition as the ground-truth source of predictive disparity.

ROME also involves several practical considerations. 
Fairness-relevant attributes must be available during model development to learn subgroup structure, which limits applicability when such attributes cannot be collected or observed. 
The two ROME formulations also present different modeling trade-offs. 
ROME-EM provides a more structured and interpretable representation of latent heterogeneity but relies on parametric mixture assumptions and can be sensitive to initialization. 
ROME-MoE relaxes these assumptions and can capture more complex nonlinear structure, but its neural gating and expert networks are less interpretable. 
This may limit its use in applications where transparent explanations of subgroup assignment or individual predictions are required.
Developing more interpretable nonlinear formulations and explanation methods for latent-group robust prediction is an important direction for future work.

\section*{Acknowledgments}
 S.L. was supported by the Khoo Postdoctoral Fellowship Award (Duke-NUS-KPFA/2025/0081), funded by the Estate of Tan Sri Khoo Teck Puat.
 The  funders had no role in study design, data collection, analysis, interpretation, or manuscript preparation.

\section*{Data and Code Availability}
Code implementing ROME is available at \href{https://github.com/siqili0325/ROME}{https://github.com/siqili0325/ROME}.
All datasets used in this study are publicly available. 
The Law School Admissions and Communities \& Crime datasets are obtained from their original
public sources, and the ACS PUMS data are accessed through the Folktables package~\citep{ding2021retiring}.

\bibliography{iclr2026_conference}
\bibliographystyle{iclr2026_conference}

\newpage
\appendix

\section{Simulation Study: Additional Details}
\label{app:sim_para}
\subsection{Parameter Specifications}\label{syn_details}
We follow~\eqref{formula:group-specific} to generate the data, using the model parameters as detailed in Section \ref{simulations}.
Specifically, we simulate features \(X=(A,S)\in\mathbb{R}^p\) by drawing \(X\sim\mathcal{N}(0,\Sigma_X)\). 
Outcomes are generated by group-specific linear models: \(Y_i=\beta_{G_i}^\top[1, A_i, S^{\text{out}}_i]+\varepsilon_i\) with \(\varepsilon_i\sim\mathcal{N}(0,0.1^2)\).
To model mixture shift, we optionally use distinct membership parameters \(\Gamma\) (train) and $\Gamma$'(test). 
Finally, to mimic imperfect initialization, each true group label is flipped to a different class with small probability $r$ to form $G_0^{\text{fake}}$. 

Full details and implementations are provided in script `demo.R' of the public code repository.

\textbf{Group Membership Parameters ($\gamma$):}
The membership parameter matrix $\gamma \in \mathbb{R}^{4 \times 5}$ was specified as:
\begin{align}
\gamma = \begin{bmatrix}
2.0 & 2.0 & 2.0 & 2.0 & 2.0 \\
-3.0 & -2.0 & -5.0 & 0.1 & 0.1 \\
0.1 & -10.0 & 0.1 & 0.1 & 0.1 \\
-2.0 & -2.0 & -2.0 & -2.0 & -2.0
\end{bmatrix}
\end{align}

\textbf{Outcome Model Coefficients ($\boldsymbol{\beta}$):}
The true regression coefficients were generated using a heterogeneous scaling approach. Starting with a base vector $\boldsymbol{\beta}_0 = \boldsymbol{1}_{20}$, group-specific coefficients were created as $\boldsymbol{\beta}_j = \epsilon_j \odot \boldsymbol{\beta}_0$, where $\epsilon_j$ are scaling factors generated with heterogeneity parameter $\delta = 0.8$. 
Group intercepts were drawn uniformly from $[-1, 1]$. 
The final coefficient matrix $\boldsymbol{\beta}^T$ (transposed for space) is:
{\small
\begin{align}
\boldsymbol{\beta}^T = \begin{bmatrix}
0.844 & 0.090 & 0.962 & 0.618 \\
-0.423 & 0.749 & 1.309 & 0.307 \\
0.696 & -0.545 & 0.559 & 1.703 \\
-0.449 & 1.646 & -1.165 & 1.361 \\
-0.737 & -0.429 & 0.255 & 1.384 \\
1.144 & -1.003 & 1.014 & -1.377 \\
0.988 & 1.666 & -1.336 & -1.209 \\
-1.702 & -1.681 & -1.295 & 0.745 \\
1.217 & 1.800 & -1.767 & 0.218 \\
-0.922 & -1.566 & 0.744 & -0.287 \\
0.403 & -1.523 & 0.395 & -1.727 \\
1.729 & 1.151 & 0.292 & 0.830 \\
-1.661 & 1.461 & 1.628 & -1.368 \\
-0.217 & 1.637 & 0.840 & -1.781 \\
-0.437 & -0.831 & -0.509 & 1.747 \\
-1.520 & -0.487 & -0.325 & -1.428 \\
-0.372 & -0.557 & 1.058 & 0.414 \\
0.826 & -1.509 & -0.450 & 0.903 \\
-1.453 & -0.848 & -0.748 & 1.429 \\
-0.773 & 0.996 & 0.765 & 1.460 \\
-1.134 & -1.441 & 0.509 & -1.790
\end{bmatrix}
\end{align}
}
\newline
where rows 1-21 correspond to the intercept, $A_1$-$A_{15}$, and $S_1$-$S_5$ respectively, and columns 1-4 correspond to latent groups 1-4.

\textbf{Algorithm Hyperparameters:}
\begin{itemize}
    \item EM convergence tolerance: $\tau_1 = 10^{-3}$ (parameter change), $\tau_2 = 5 \times 10^{-3}$ (line search)
    \item Maximum EM iterations: 100
    \item DRO constraint values: $\mathcal{C} \in \{1.0, 0.6, 0.5, 0.49, 0.48, 0.47, \ldots, 0.04, 0.03, 0.02\}$ (27 values total)
    \item Misspecification rate: 50\% (proportion of observations with incorrect initial group assignment)
\end{itemize}

\textbf{Data Structure:}
All features were generated from independent standard normal distributions ($\Sigma_X = \boldsymbol{I}_{20}$). The first 15 features served as non-sensitive features ($\boldsymbol{A}$), while the remaining 5 were treated as sensitive attributes ($\boldsymbol{S}$). 
All five sensitive attributes were included in both the membership ($\mathcal{I}_{\text{mem}} = \{1,2,3,4,5\}$) and outcome ($\mathcal{I}_{\text{out}} = \{1,2,3,4,5\}$) models in this simulation.

\subsection{Additional Simulation Results}
\label{app:additional_groups}

\begin{figure}[!htbp]
    \centering
    \includegraphics[width=1\linewidth]{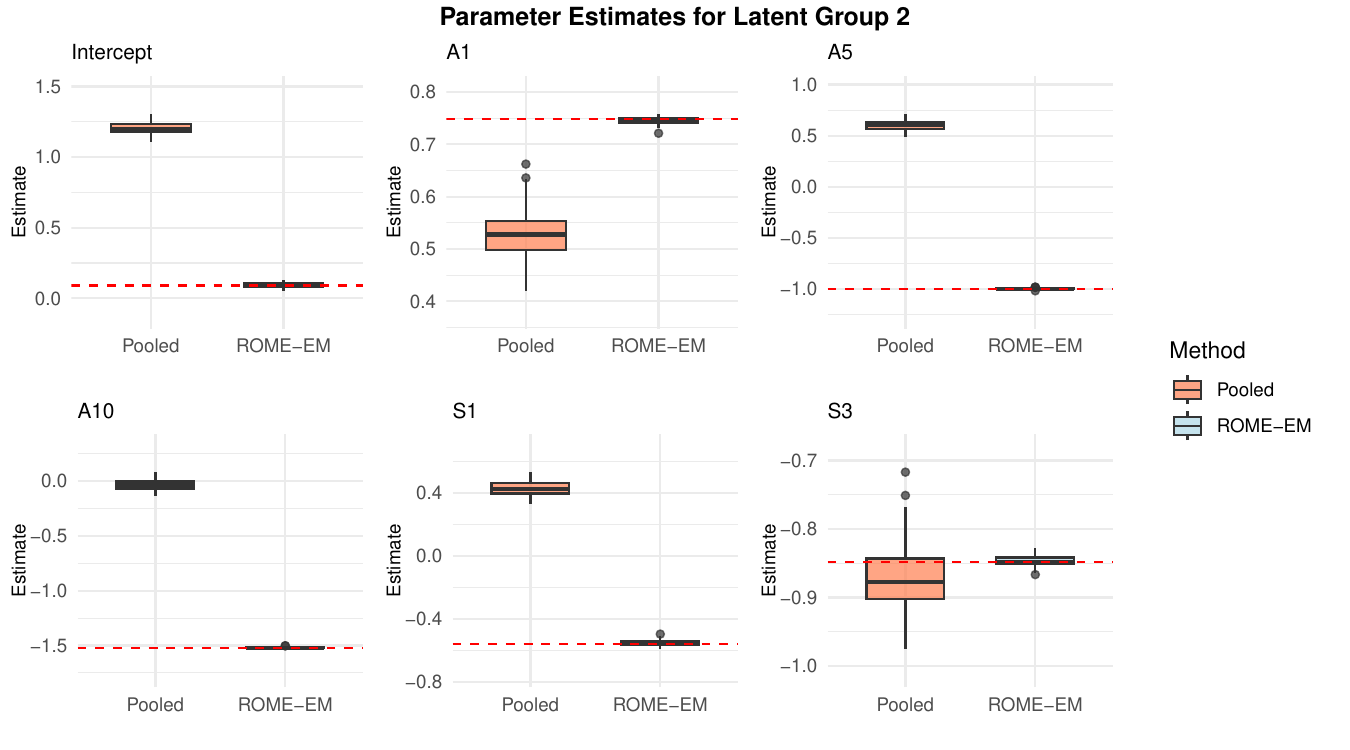}
    \caption{Box plots comparing parameter recovery by ROME-EM and pooled regression for latent group 2 (over 100 simulations). Red dashed lines denote ground truth values for each parameter.}
    \label{fig:bias_group2}
\end{figure}

\begin{figure}[!htbp]
    \centering
    \includegraphics[width=1\linewidth]{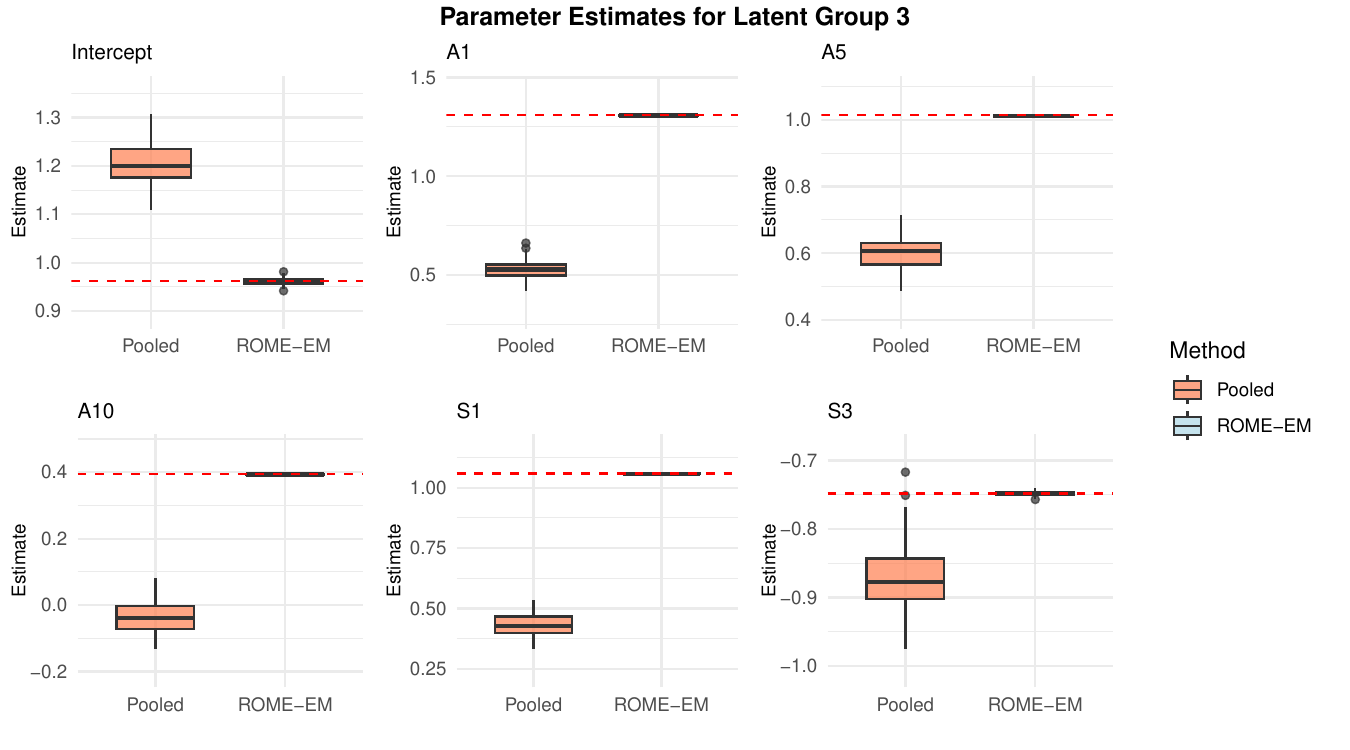}
    \caption{Box plots comparing parameter recovery by ROME-EM and pooled regression for latent group 3 (over 100 simulations). Red dashed lines denote ground truth values for each parameter.}
    \label{fig:bias_group3}
\end{figure}

\begin{figure}[!htbp]
    \centering
    \includegraphics[width=1\linewidth]{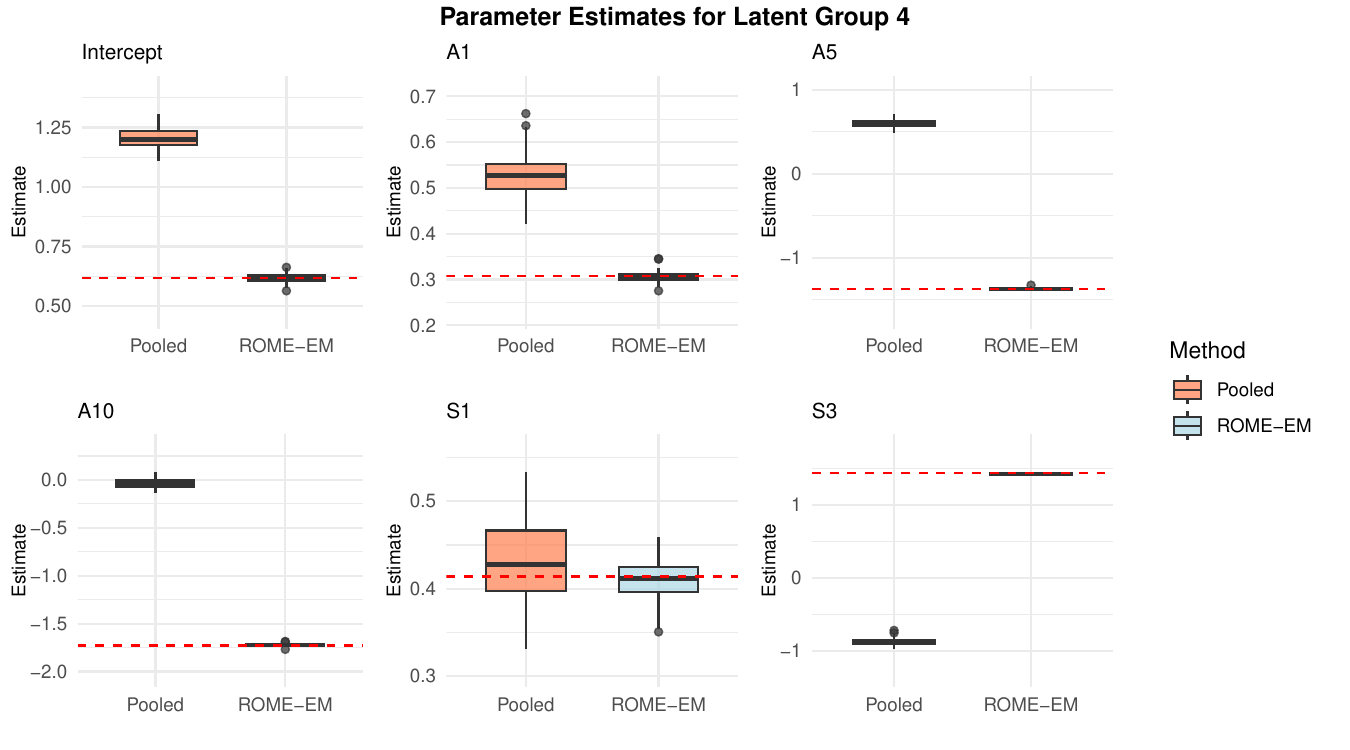}
    \caption{Box plots comparing parameter recovery by ROME-EM and pooled regression for latent group 4 (over 100 simulations). Red dashed lines denote ground truth values for each parameter.}
    \label{fig:bias_group4}
\end{figure}

Figures \ref{fig:bias_group2}, \ref{fig:bias_group3}, and \ref{fig:bias_group4} present the additional simulation results as complementary to Figure \ref{fig:bias_group1} in the main text.

\section{Details of Real-datasets}\label{realdatadetails}

\paragraph{Law School Admissions Council~\citep{wightman1999lsac}}
We use `zfygpa' as $Y$, features `race1\_black' ($S_1$), `gender\_male' ($S_2$) and `age' ($S_3$) as $\boldsymbol{S}$; `lsat',`ugpa',`fam\_inc', `tier' and `fulltime' as $\boldsymbol{A}$.

\paragraph{Communities and Crime~\citep{communities_and_crime_183}}
We use `ViolentCrimesPerPop' as $Y$, features `racepctblack' ($S_1$), `racePctHisp' ($S_2$) and `agePct12t21' ($S_3$) as $\boldsymbol{S}$; and features `PctUnemployed', `medIncome', `PopDens', `PolicPerPop', `MedRent', `PctFam2Par',`PctIlleg', `LandArea' and `pctWWage' as $\boldsymbol{A}$.

\paragraph{American Community Survey Public Use Microdata Sample~\citep{us2003american}} 
We use `PINCP' as $Y$, features  `SEX' ($S_1$), `RAC1P'($S_2$), `AGEP' ($S_3$) and `SCHL' ($S_4$) as $\boldsymbol{S}$; and features `COW', `MAR', `OCCP', `POBP', `RELP', `WKHP' as $\boldsymbol{A}$.

\section{Ablation Studies}

\subsection{Impact of DRO Parameter}\label{append.ablation.alpha}

\begin{figure}[!htbp]
    \centering
    \includegraphics[width=0.9\linewidth]{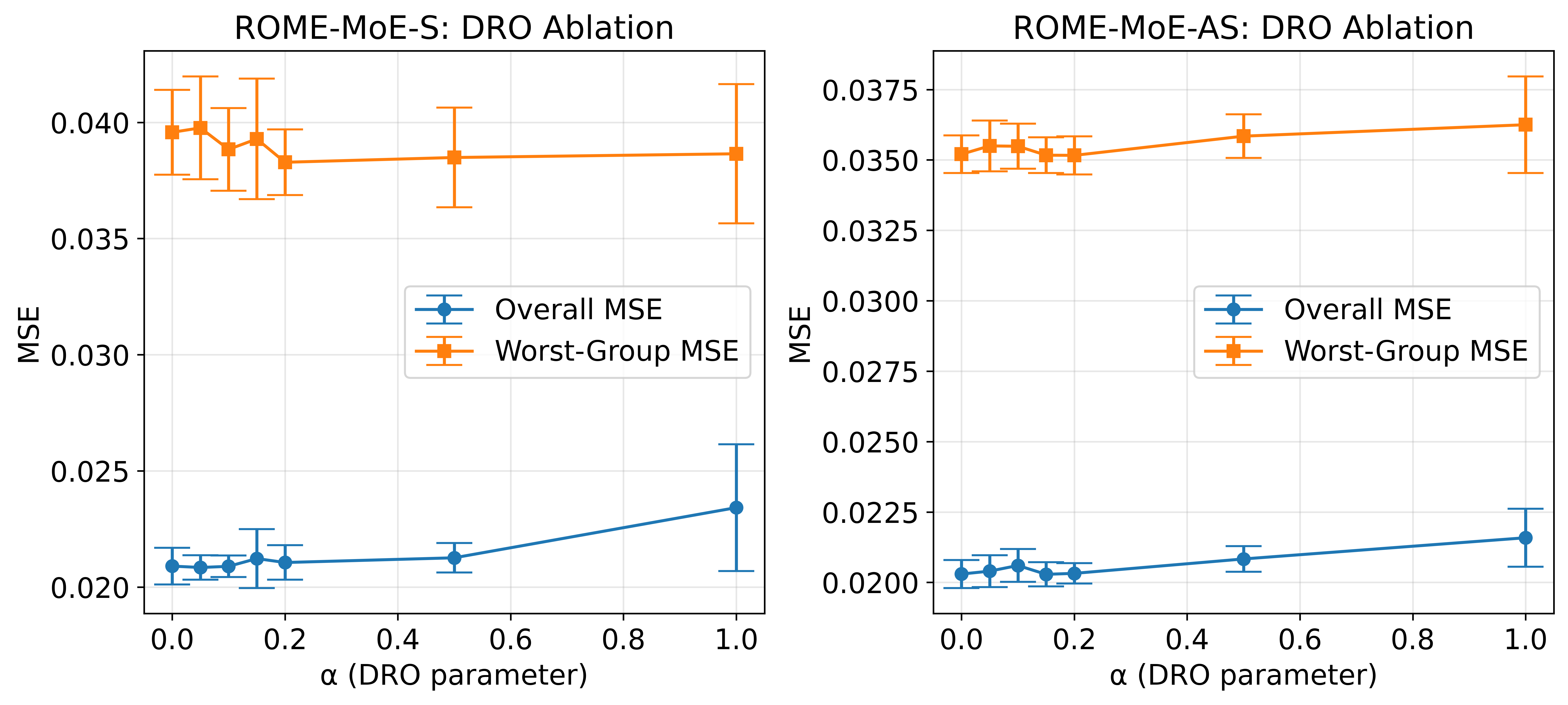}
    \caption{Effect of DRO parameter $\alpha$ on ROME-MoE performance (Communities \& Crime dataset, 10 seeds). Both variants show worst-group improvements with minimal overall performance degradation as $\alpha$ increases from 0 (standard training) to 1 (pure worst-group optimization).}
    \label{crime}
\end{figure}

\begin{figure}[!htbp]
    \centering
    \includegraphics[width=0.9\linewidth]{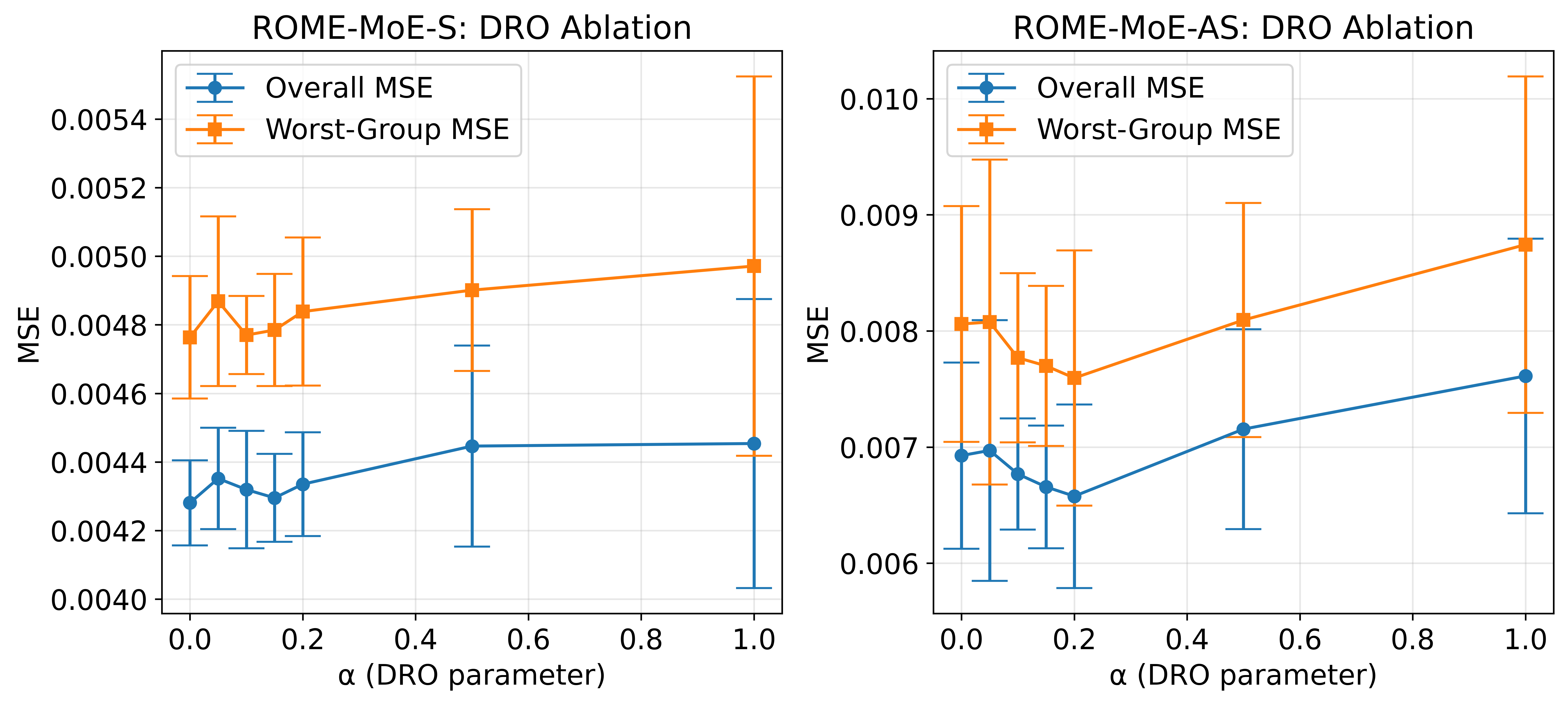}
    \caption{Effect of DRO parameter $\alpha$ on ROME-MoE performance (American Community Survey dataset, 10 seeds). Both variants show worst-group improvements with minimal overall performance degradation as $\alpha$ increases from 0 (standard training) to 1 (pure worst-group optimization).}
    \label{acs}
\end{figure}

Figures \ref{crime} and \ref{acs} present the additional results for $\alpha$ ablation studies as complementary to Figure \ref{law} in the main text.

\subsection{Impact of Evaluation Subgroups}\label{append.ablation.subgroups}
Evaluation subgroups are formed by partitioning sensitive attributes: categorical variables retain their discrete values, while continuous variables are discretized using median or quartile splits.
Besides Tables \ref{tab:main_results_mse} and \ref{tab:main_results_r2} in main text, where we use quartile cuts for ($S_2, S_3$, quartile) for law dataset, ($S_2, S_3$, median)  for crime dataset and $S_2$ for American Community Survey dataset, where feature details are available in \ref{realdatadetails}, we conduct ablation studies by further using different evaluation subgroups.

\begin{table}[!htbp]
\centering
\caption{Ablation study on evaluation subgroup schemes for the Law School dataset. 
Results show worst-group performance across different sensitive attribute combinations, averaged over 10 random seeds (mean $\pm$ standard error). 
Best-performing fair models are shown in \textbf{bold}.
Statistical significance is assessed using paired two-sided t-tests across the 10 matched random seeds, comparing each ROME variant with the best-performing fair baseline for each evaluation subgroup scheme: $^*p<0.05$, $^{**}p<0.01$, and $^{***}p<0.001$; $^{}$ indicates no significant difference ($p>0.05$).}
\label{tab:ablation_law}
\begin{tabular}{@{}llcll@{}}
\toprule
\textbf{Evaluation Subgroups} & \textbf{Model} & \textbf{Fair}  & \textbf{Worst-Group MSE} & \textbf{Worst-Group $\boldsymbol{R^2}$}\\ \midrule
\multirow{3}{*}{\begin{tabular}[c]{@{}l@{}} $S_1$, $S_3$, median \end{tabular}} 
 & Baseline MLP &  & 0.7640 $\pm$ 0.0007 & 0.1729 $\pm$ 0.0656\\
 & Baseline MLP - Fair & \checkmark & 0.7795 $\pm$ 0.0047 & 0.0402 $\pm$ 0.0827 \\

& ARL & \checkmark & 0.7744 $\pm$ 0.0034 & -0.0202 $\pm$ 0.0102 \\

& FairReg-BGL & \checkmark & 0.7730 $\pm$ 0.0024 & -0.0227 $\pm$ 0.0098 \\

 & Vanilla MoE &  & 0.7656 $\pm$ 0.0010 & 
0.1719 $\pm$ 0.0663 \\
 & ROME-MoE-S & \checkmark & 0.7628 $\pm$ 0.0010$^{**}$ & 
\textbf{0.1710 $\pm$ 0.0664}$^{***}$ \\
 & ROME-MoE-AS & \checkmark & \textbf{0.7622 $\pm$ 0.0015}$^{**}$ & 
 0.1581 $\pm$ 0.0689$^{**}$ \\
 \midrule
 
\multirow{3}{*}{$S_2$, $S_3$, median} 
 & Baseline MLP &  & 0.7674 $\pm$ 0.0010 & 0.1276 $\pm$ 0.0006 \\
 & Baseline MLP - Fair & \checkmark & 0.7825 $\pm$ 0.0015 & 0.1029 $\pm$ 0.0011 \\

& ARL & \checkmark & 0.7811 $\pm$ 0.0013 & 0.1037 $\pm$ 0.0008 \\

& FairReg-BGL & \checkmark & 0.7787 $\pm$ 0.0016 & 0.1010 $\pm$ 0.0007 \\
 
 & Vanilla MoE &  & 0.7681 $\pm$ 0.0015 & 
 0.1277 $\pm$ 0.0006 \\
 & ROME-MoE-S & \checkmark & 0.7675 $\pm$ 0.0019$^{***}$ & 
 0.1210 $\pm$ 0.0020$^{***}$ \\
 & ROME-MoE-AS & \checkmark & \textbf{0.7644 $\pm$ 0.0016}$^{***}$ & \textbf{0.1235 $\pm$ 0.0012}$^{***}$ \\
 \bottomrule
\end{tabular}
\end{table}

\begin{table}[!htbp]
\centering
\caption{Ablation study on evaluation subgroup schemes for the Communities \& Crime dataset. Results show worst-group performance across different sensitive attribute combinations, averaged over 10 random seeds (mean $\pm$ standard error). 
Best-performing fair models are shown in \textbf{bold}. 
Statistical significance is assessed using paired two-sided t-tests across the 10 matched random seeds, comparing each ROME variant with the best-performing fair baseline for each metric and evaluation subgroup scheme: $^*p<0.05$, $^{**}p<0.01$, and $^{***}p<0.001$; $^{}$ indicates no significant difference ($p>0.05$).}
\label{tab:ablation_crime}
\begin{tabular}{@{}llcll@{}}
\toprule
\textbf{Evaluation Subgroups} & \textbf{Model} & \textbf{Fair}  & \textbf{Worst-Group MSE} & \textbf{Worst-Group $\boldsymbol{R^2}$}\\ \midrule
\multirow{3}{*}{\begin{tabular}[c]{@{}l@{}} $S_1$, $S_2$, median \end{tabular}} 
 & Baseline MLP &  & 0.0402 $\pm$ 0.0005 & 0.2106 $\pm$ 0.0081\\
 & Baseline MLP - Fair & \checkmark & 0.0418 $\pm$ 0.0003 & -0.0542 $\pm$ 0.0425 \\

& ARL & \checkmark & 0.0381 $\pm$ 0.0003 & -0.0085 $\pm$ 0.0243 \\ 

& FairReg-BGL & \checkmark & 0.0379 $\pm$ 0.0002 & 0.0032 $\pm$ 0.0104 \\

 & Vanilla MoE &  & 0.0404 $\pm$ 0.0007 & 0.2100 $\pm$ 0.0084  \\
 & ROME-MoE-S & \checkmark & 0.0407 $\pm$ 0.0005 & 
\textbf{0.2005 $\pm$ 0.0061}$^{***}$ \\
 & ROME-MoE-AS & \checkmark & \textbf{0.0369 $\pm$ 0.0002}$^{*}$ & 
 0.1406 $\pm$ 0.0108$^{***}$ \\
 \midrule
 
\multirow{3}{*}{$S_1$, $S_3$, median} 
 & Baseline MLP &  & 0.0332 $\pm$ 0.0004 & 0.1847 $\pm$ 0.0076 \\
 & Baseline MLP - Fair & \checkmark & 0.0385 $\pm$ 0.0004 & 0.1352 $\pm$ 0.0156 \\

 & ARL & \checkmark & 0.0362 $\pm$ 0.0003 & 0.1322 $\pm$ 0.0082 \\

 & FairReg-BGL & \checkmark & 0.0360 $\pm$ 0.0002 & 0.1405 $\pm$ 0.0028 \\
 
 & Vanilla MoE &  & 0.0339 $\pm$ 0.0007 & 
0.1885 $\pm$ 0.0079 \\
 & ROME-MoE-S & \checkmark & 0.0339 $\pm$ 0.0004$^{***}$ & 
 \textbf{0.1796 $\pm$ 0.0051}$^{***}$ \\
 & ROME-MoE-AS & \checkmark & \textbf{0.0319 $\pm$ 0.0002}$^{***}$ & 0.1520 $\pm$ 0.0050 \\
 \bottomrule
\end{tabular}
\end{table}

\begin{table}[!htbp]
\centering
\caption{Ablation study on evaluation subgroup schemes for the American Community Survey Public Use Microdata Sample dataset. 
Results show worst-group performance across different sensitive attribute combinations, averaged over 10 random seeds (mean $\pm$ standard error). 
Best-performing fair models are shown in \textbf{bold}. 
Statistical significance is assessed using paired two-sided t-tests across the 10 matched random seeds, comparing each ROME variant with the best-performing fair baseline for each metric and evaluation subgroup scheme: $^*p<0.05$, $^{**}p<0.01$, and $^{***}p<0.001$; $^{}$ indicates no significant difference ($p>0.05$).}
\label{tab:ablation_acs}
\begin{tabular}{@{}llcll@{}}
\toprule
\textbf{Evaluation Subgroups} & \textbf{Model} & \textbf{Fair}  & \textbf{Worst-Group MSE} & \textbf{Worst-Group $\boldsymbol{R^2}$}\\ \midrule

\multirow{3}{*}{\begin{tabular}[c]{@{}l@{}} $S_1$ \end{tabular}} 
 & Baseline MLP &  & 0.0047 $\pm$ 0.0000 & 0.4870 $\pm$ 0.0037\\
 & Baseline MLP - Fair & \checkmark & 0.0055 $\pm$ 0.0001 & 0.4026 $\pm$ 0.0054  \\

  & ARL & \checkmark & 0.0056 $\pm$ 0.0000 & 0.4008 $\pm$ 0.0040 \\

  & FairReg-BGL & \checkmark & 0.0056 $\pm$ 0.0001 & 0.3959 $\pm$ 0.0071 \\
  
 & Vanilla MoE &  & 0.0052 $\pm$ 0.0000 & 0.4451 $\pm$ 0.0038  \\
 & ROME-MoE-S & \checkmark & \textbf{0.0049 $\pm$ 0.0001}$^{***}$ & \textbf{0.4678 $\pm$ 0.0060}$^{***}$ 
 \\
 & ROME-MoE-AS & \checkmark & 0.0051 $\pm$ 0.0001$^{***}$ & 0.4499 $\pm$ 0.0065$^{***}$
  \\
 \midrule

\multirow{3}{*}{\begin{tabular}[c]{@{}l@{}} $S_1$ and $S_2$  \end{tabular}} 
 & Baseline MLP &  & 0.0049 $\pm$ 0.0000  & 0.4445 $\pm$ 0.0053\\
 & Baseline MLP - Fair & \checkmark &  0.0056 $\pm$ 0.0001 &  0.3764 $\pm$ 0.0067\\

& ARL & \checkmark & 0.0056 $\pm$ 0.0000 & 0.3827 $\pm$ 0.0067 \\

& FairReg-BGL & \checkmark & 0.0057 $\pm$ 0.0001 & 0.3822 $\pm$ 0.0067 \\

 & Vanilla MoE &  & 0.0052 $\pm$ 0.0000 & 0.4039 $\pm$ 0.0049  \\
 & ROME-MoE-S & \checkmark &  \textbf{0.0050 $\pm$ 0.0001}$^{***}$  & \textbf{0.4328 $\pm$ 0.0080}$^{**}$
 \\
 & ROME-MoE-AS & \checkmark & 0.0052 $\pm$ 0.0001$^{***}$  & 0.4074 $\pm$ 0.0080
  \\
 \midrule

 \bottomrule
\end{tabular}
\end{table}

\section{Neural Network Details}

\subsection{Neural Network  Architecture}
All MoE-based models (Vanilla MoE, ROME-MoE-S, ROME-MoE-AS) use standard two-layer multilayer perceptrons (MLPs) for both the experts and the gating network. 
Each expert has the structure \texttt{Linear–ReLU–Linear} and outputs a scalar prediction. 
The Vanilla MoE experts take $(A,S)$ as input, while the fair variants (ROME-MoE-S / ROME-MoE-AS) use only $A$ as expert input. 
The gating network is also a two-layer MLP, taking either $S$ (ROME-MoE-S) or $(A,S)$ (ROME-MoE-AS) as input and producing $G$ logits, followed by a softmax to obtain mixture weights. 
The final prediction is the weighted combination of expert outputs.
Hyperparameter search spaces and selected values (e.g., hidden sizes $\{32, 64, 128\}$, number of experts $G \in \{2,3,4\}$) are provided in Tables~\ref{tab:hyperparams_law}, \ref{tab:hyperparams_crime} and \ref{tab:hyperparams_acs}.

All baseline MLP models use a standard two-layer multilayer perceptron (MLP) of the form Linear $\rightarrow$ ReLU $\rightarrow$ Linear, producing a scalar prediction.
The variants differ only in their input features and training objective: Baseline MLP–Full (unfair): experts receive the full feature vector (A, S); Baseline MLP–Fair: experts receive only A, matching the fairness constraint used in ROME-MoE.
Hyperparameter search spaces (hidden sizes \{32, 64, 128\}, learning rates, etc.) are identical to those used for MoE and are summarized in Tables \ref{tab:hyperparams_law}–\ref{tab:hyperparams_acs}.

\paragraph{Fair Regression-BGL.}
We implement the bounded group loss (BGL) formulation of Fair Regression~\citep{pmlr-v97-agarwal19d} using the
\texttt{BoundedGroupLoss} and \texttt{ExponentiatedGradient} reductions in Fairlearn.
BGL minimizes overall prediction loss subject to an upper bound $\zeta$ on the loss within each predefined sensitive group. 
We use squared loss and an $\boldsymbol{A}$-only Linear--ReLU--Linear MLP as the base regression learner, with hidden sizes 32, 64, and 32 for Law School, Communities \& Crime, and ACS PUMS, respectively. 
We fix the learning rate at $10^{-3}$, use a batch size of 64, and train for 30 epochs. 
Sensitive attributes are used to construct the BGL constraints but are not provided as inputs to the regression learner.

For each dataset, BGL is trained using the sensitive-group definition corresponding to the main evaluation setting: $(S_2,S_3)$ with quartile discretization for Law School, $(S_2,S_3)$ with median discretization for Communities \& Crime, and $S_2$ for ACS PUMS. 
We first perform a coarse search over the BGL upper bound $\zeta$ and then refine the search around the feasible region using the held-out validation split. 
Among approximately feasible configurations, we select $\zeta$ according to worst-group validation MSE, using overall validation MSE as a tie-breaker. 
The final refined search spaces and selected values are reported in Tables~\ref{tab:hyperparams_law}, \ref{tab:hyperparams_crime}, and~\ref{tab:hyperparams_acs}.
For evaluation, we use the deterministic weighted prediction from the mixture of predictors returned by the exponentiated-gradient reduction.

\subsection{Hyperparameter Tuning}\label{sec:appendix_hyperparams}
Hyperparameters were selected using a held-out 20\% subset of the original training data, with the test set held out throughout model selection. 
The search spaces and final selected values for each dataset are reported in Tables~\ref{tab:hyperparams_law}, \ref{tab:hyperparams_crime}, and~\ref{tab:hyperparams_acs}.
For ARL, we tune the learner and adversary learning rates while fixing the learner architecture. 
For Fair Regression-BGL, we fix the base learner and tune the BGL upper bound $\zeta$. 
After hyperparameter selection, models are refit using the full training set and evaluated on the held-out test set over 10 random seeds.

\begin{table}[!htbp]
\centering
\caption{Hyperparameter tuning results for Law School Admissions Council}
\label{tab:hyperparams_law}
\begin{tabular}{@{}lcc@{}}
\toprule
\textbf{Hyperparameter} & \textbf{Search Space} & \textbf{Best Value} \\ \midrule
\multicolumn{3}{l}{\textit{BaselineMLP}} \\
\quad Learning Rate & \{1e-2, 1e-3, 1e-4\} & 1e-4 \\
\quad Hidden Size & \{32, 64, 128\} & 64 \\
\midrule
\multicolumn{3}{l}{\textit{BaselineMLP-Fair}} \\
\quad Learning Rate & \{1e-2, 1e-3, 1e-4\} & 1e-3 \\
\quad Hidden Size & \{32, 64, 128\} & 32 \\
\midrule

\multicolumn{3}{l}{\textit{ARL}} \\
\quad Learner Learning Rate
& \{1e-3, 1e-4\} & 1e-3 \\
\quad Adversary Learning Rate
& \{1e-3, 3e-4, 1e-4, 3e-5\} & 1e-4 \\
\quad Hidden Size
& 32 & 32 \\

\midrule
\multicolumn{3}{l}{\textit{Fair Regression-BGL}} \\
\quad BGL Upper Bound $\zeta$
& \{0.72, 0.74, 0.76, 0.78, 0.80\}
& 0.78 \\

\midrule

\multicolumn{3}{l}{\textit{Vanilla MoE}} \\
\quad Learning Rate & \{1e-2, 1e-3, 1e-4\} & 1e-4 \\
\quad Hidden Size (expert) & \{32, 64, 128\} & 32 \\
\quad Hidden Size (gating) & \{32, 64, 128\} & 64 \\
\quad Number of Experts & \{2, 3, 4\} & 4 \\
\midrule
\multicolumn{3}{l}{\textit{ROME-MoE-S}} \\
\quad Learning Rate & \{1e-2, 1e-3, 1e-4\} & 1e-4 \\
\quad Hidden Size (expert) & \{32, 64, 128\} & 128 \\
\quad Hidden Size (gating) & \{32, 64, 128\} & 16 \\
\quad Number of Experts & \{2, 3, 4\} & 2 \\
\midrule
\multicolumn{3}{l}{\textit{ROME-MoE-AS}} \\
\quad Learning Rate & \{1e-2, 1e-3, 1e-4\} & 1e-4 \\
\quad Hidden Size (expert) & \{32, 64, 128\} & 64 \\
\quad Hidden Size (gating) & \{32, 64, 128\} & 64 \\
\quad Number of Experts & \{2, 3, 4\} & 2 \\
\bottomrule
\end{tabular}
\end{table}

\begin{table}[!htbp]
\centering
\caption{Hyperparameter tuning results for Communities and Crime dataset}
\label{tab:hyperparams_crime}
\begin{tabular}{@{}lcc@{}}
\toprule
\textbf{Hyperparameter} & \textbf{Search Space} & \textbf{Best Value} \\ \midrule
\multicolumn{3}{l}{\textit{BaselineMLP}} \\
\quad Learning Rate & \{1e-2, 1e-3, 1e-4\} & 1e-2 \\
\quad Hidden Size & \{32, 64, 128\} & 128 \\
\midrule
\multicolumn{3}{l}{\textit{BaselineMLP-Fair}} \\
\quad Learning Rate & \{1e-2, 1e-3, 1e-4\} & 1e-2 \\
\quad Hidden Size & \{32, 64, 128\} & 64 \\
\midrule

\multicolumn{3}{l}{\textit{ARL}} \\
\quad Learner Learning Rate
& \{1e-3, 1e-4\} & 1e-3 \\
\quad Adversary Learning Rate
& \{1e-3, 3e-4, 1e-4, 3e-5\} & 3e-4 \\
\quad Hidden Size
& 64 & 64 \\

\midrule
\multicolumn{3}{l}{\textit{Fair Regression-BGL}} \\
\quad BGL Upper Bound $\zeta$
& \{0.0260, 0.0265, 0.0270, 0.0275, 0.0280\}
& 0.0270 \\

\midrule
\multicolumn{3}{l}{\textit{Vanilla MoE}} \\
\quad Learning Rate & \{1e-2, 1e-3, 1e-4\} & 1e-2 \\
\quad Hidden Size (expert) & \{32, 64, 128\} & 64 \\
\quad Hidden Size (gating) & \{32, 64, 128\} & 32 \\
\quad Number of Experts & \{2, 3, 4\} & 2 \\
\midrule
\multicolumn{3}{l}{\textit{ROME-MoE-S}} \\
\quad Learning Rate & \{1e-2, 1e-3, 1e-4\} & 1e-2 \\
\quad Hidden Size (expert) & \{32, 64, 128\} & 64 \\
\quad Hidden Size (gating) & \{32, 64, 128\} & 16 \\
\quad Number of Experts & \{2, 3, 4\} & 3 \\
\midrule
\multicolumn{3}{l}{\textit{ROME-MoE-AS}} \\
\quad Learning Rate & \{1e-2, 1e-3, 1e-4\} & 1e-3 \\
\quad Hidden Size (expert) & \{32, 64, 128\} & 32 \\
\quad Hidden Size (gating) & \{32, 64, 128\} & 16 \\
\quad Number of Experts & \{2, 3, 4\} & 2 \\
\bottomrule
\end{tabular}
\end{table}

\begin{table}[!htbp]
\centering
\caption{Hyperparameter tuning results for American Community Survey Public Use Microdata Sample dataset}
\label{tab:hyperparams_acs}
\begin{tabular}{@{}lcc@{}}
\toprule
\textbf{Hyperparameter} & \textbf{Search Space} & \textbf{Best Value} \\ \midrule
\multicolumn{3}{l}{\textit{BaselineMLP}} \\
\quad Learning Rate & \{1e-2, 1e-3, 1e-4\} & 1e-3 \\
\quad Hidden Size & \{32, 64, 128\} & 64\\
\midrule
\multicolumn{3}{l}{\textit{BaselineMLP-Fair}} \\
\quad Learning Rate & \{1e-2, 1e-3, 1e-4\} & 1e-3 \\
\quad Hidden Size & \{32, 64, 128\} & 32 \\
\midrule

\multicolumn{3}{l}{\textit{ARL}} \\
\quad Learner Learning Rate
& \{1e-3, 1e-4\} & 1e-3 \\
\quad Adversary Learning Rate
& \{1e-3, 3e-4, 1e-4, 3e-5\} & 1e-4 \\
\quad Hidden Size
& 32 & 32 \\

\midrule
\multicolumn{3}{l}{\textit{Fair Regression-BGL}} \\
\quad BGL Upper Bound $\zeta$
& \{0.00520, 0.00530, 0.00540, 0.00545, 0.00550\}
& 0.00550 \\

\midrule
\multicolumn{3}{l}{\textit{Vanilla MoE}} \\
\quad Learning Rate & \{1e-2, 1e-3, 1e-4\} & 1e-4 \\
\quad Hidden Size (expert) & \{32, 64, 128\} & 128 \\
\quad Hidden Size (gating) & \{32, 64, 128\} & 16 \\
\quad Number of Experts & \{2, 3, 4\} & 4 \\
\midrule
\multicolumn{3}{l}{\textit{ROME-MoE-S}} \\
\quad Learning Rate & \{1e-2, 1e-3, 1e-4\} & 1e-2 \\
\quad Hidden Size (expert) & \{32, 64, 128\} & 128 \\
\quad Hidden Size (gating) & \{32, 64, 128\} & 32 \\
\quad Number of Experts & \{2, 3, 4\} & 2 \\
\midrule
\multicolumn{3}{l}{\textit{ROME-MoE-AS}} \\
\quad Learning Rate & \{1e-2, 1e-3, 1e-4\} & 1e-4 \\
\quad Hidden Size (expert) & \{32, 64, 128\} & 128 \\
\quad Hidden Size (gating) & \{32, 64, 128\} & 32 \\
\quad Number of Experts & \{2, 3, 4\} & 4 \\
\bottomrule
\end{tabular}
\end{table}

\end{document}

%% file: math_commands.tex
\usepackage{amsmath,amsfonts,bm}

\def\eqref#1{equation~\ref{#1}}

\def\1{\bm{1}}

\DeclareMathAlphabet{\mathsfit}{\encodingdefault}{\sfdefault}{m}{sl}
\SetMathAlphabet{\mathsfit}{bold}{\encodingdefault}{\sfdefault}{bx}{n}

